\documentclass[a4paper]{article}

\usepackage[latin1]{inputenc}
\usepackage[english]{babel}
\usepackage[final]{graphicx}
\usepackage{verbatim}

\usepackage{amsmath,amsfonts,amssymb,amsthm}
\usepackage{latexsym}

\newcommand{\authorstructure}[1]{{\small \it #1}}

\newenvironment{Aabstract}
	{\vspace{5mm}
	 \begin{center}\bf Abstract\end{center}
	 \begin{quote}\small}
	{\end{quote}\vspace{5mm}}

\begin{document}

\begin{center}
{\LARGE Automatic Classification using Self-Organising Neural Networks in Astrophysical Experiments}
\\[5mm]
{\sc Praveen Boinee, Alessandro De Angelis, Edoardo Milotti}
\end{center}	
\authorstructure{Dipartimento di Fisica dell'Universit\`a di~Udine e INFN, Sez.~di Trieste, Gruppo Collegato di Udine, via
delle Scienze~208, 33100~Udine, Italy}

\begin{Aabstract}
Self-Organising Maps (SOMs) are effective tools in classification problems, and in recent years the even more powerful Dynamic Growing Neural Networks, a variant of SOMs, have been developed. Automatic Classification (also called {\it clustering}) is an important and difficult problem in many Astrophysical experiments, for instance, Gamma Ray Burst classification, or gamma-hadron separation.
After a brief introduction to classification problem, we discuss Self-Or\-ganis\-ing Maps in section 2. Section 3 discusses with various models of growing neural networks and finally in section 4  we discuss the research perspectives in  growing neural networks for efficient classification in astrophysical problems.
\end{Aabstract}

\section{Introduction}
Astrophysical database contain large amounts of data; one example is given by the growing number of experiments studying Gamma Ray Bursts (GRBs). Data sets can be found in several archives (see e.g.~Ref. \cite{p1}). 

With the growing number of experiments dedicated to GRBs it is essential to optimize the
techniques for the complex task of classification. Artificial Intelligence- (AI-) based pattern recognition algorithms are one possible candidate: automated linear classification of vector data into a given number (or an arbitrary number) of classes is a well established technique in the field of machine learning.
Several varieties of AI-based classifiers exist~\cite{p16}. Usually, one makes the distinction between supervised and unsupervised classifiers: the former are trained with data for which the classification is known and then used to classify raw data, while the latter attempt to find the best-fitting class structure in the input data by using some measure of merit.

Clustering is the unsupervised classification of patterns~\cite{p4} (observations, data items or feature vectors) into groups called clusters. Clustering is useful in several exploratory pattern analysis, grouping, decision making and machine learning situations including data mining, document retrieval, image segmentation and pattern classification.

Self-Organising Neural Networks~\cite{p2} are often used to cluster input data. Similar patterns are grouped by the network and are represented by a single unit. This grouping is done automatically on the basis of data correlations. Well-known examples of Self-Organising Artificial Neural Networks (ANN) used for clustering include Kohonen's self-organising maps, Self-Organising Tree Algorithm (SOTA), Growing Cell Structures (GCS).

\section {Self-Organising Maps (SOM)}

The basic idea is the following: a SOM defines a mapping from a high dimensional input data space onto a regular two-dimensional array of neurons~\cite{p2}. Every neuron $i$ of the map is associated with a $n$-dimensional reference vector $m_i = { [ { m_{i1},\ldots,m_{in} } ] }^T$ , where $n$ denotes the dimension of the input vectors. The set of reference vectors is called a {\it codebook}. The neurons of the map are connected to adjacent neurons by a neighbourhood relation, which dictates the topology, or the structure, of the map. The most common topologies in use are rectangular and hexagonal.
The network topology is defined by the set $N_i$ of the nearest-neighbors of neuron $i$: in the basic SOM algorithm, the topology and the total number of neurons remain fixed. The total number of neurons determines the granularity of the mapping, which has an effect on the resolution and generalization ability of the SOM.
During the training phase, the SOM acts like an elastic net that wraps the ``cloud'' formed by input data. The algorithm controls the net as it tries to approximate the density of the data, and the reference vectors in the codebook drift to the areas where the density of the input data is high. Eventually, only few codebook vectors lie in areas where the input data is sparse.

The learning process of the SOM goes as follows: 
\begin{enumerate}
\item
One sample vector $x$ is randomly drawn from the input data set and its similarity (distance) to the codebook vectors is computed by using e.g. the common Euclidean distance measure:
	$$ 
	\left\| x- m_c \right\| = 
	\min_i \left\{ \left\| x- m_i \right\| \right\} 
	$$ 

\item
After the Best Matching Unit (BMU) has been found, i.e.~the codebook vector closest to the random input vector, the codebook vectors themselves are updated. The BMU itself as well as its topological neighbours are moved closer to the input vector in the input space i.e.~the input vector attracts them. The magnitude of the attraction is governed by the learning rate. As the learning proceeds and new input vectors are given to the map, the learning rate gradually decreases to zero according to the specified learning rate function type. Along with the learning rate, the neighbourhood radius decreases as well.
If the neighborhood of the codebook vector closest to the input data vector $x(t)$ at step $t$ is $N_c(t)$, then the reference vector update rule is the following:
$$ 
m_i(t+1)=\left\{ 
\begin{alignedat}{2} 
&m_i(t) + \alpha(t)[x(t)-m_i(t)], & & \qquad i \in N_c(t) 
\\ 
&m_i(t), & & \qquad i \notin N_c(t) 
\end{alignedat} 
\right. 
$$

\item 
Steps 1 and 2 together constitute a single training step and they are repeated until the training ends. The number of training steps must be fixed prior to training the SOM because the rate of convergence in the neighbourhood function and the learning rate is calculated accordingly. 
After the training is over, the map should be topologically ordered: this means that input data vectors that are close to each other in input space map onto neurons that are close to each other in the SOM.
\end{enumerate}

Possible problems with the SOM are the following: 
\begin{itemize}
\item 
Specifying in advance a suitable size of the SOM  can be difficult.
\item 
The predefined (usually rectangular) structure may not be suitable to represent the given data.
\end{itemize}

\section{Growing Neural Networks }

Growing self-organizing networks are vector-based models trained by competitive learning~\cite{p3}. The general strategy is to start with a near-minimal network and add units until the network has the desired size or is ``good enough'' for the task at hand. This approach can be characterized by: 
\begin{itemize}
\item
generation of the network structure through a growth process;
\item
use of local statistical quantities to guide insertions and deletions of units. 
\end{itemize}
Different constraints imposed on the network structure lead to different models:
\begin{itemize}
\item	
general graph: Growing Neural Gas  (GNG);
\item
$k$-dimensional hypertetrahedrons: Growing Cell Structures (GCS);
\item
Binary Tree Structure: Self-Organising Tree Algorithm (SOTA).
\end{itemize}
Advantages over other approaches include: 
\begin{itemize}
\item
only constant parameters;
\item
network structure and size has not to be defined in advance;
\item
fast training, in particular in supervised learning (a factor of 30-100 over backprop has been observed for some problems). 
\end{itemize}
\subsection{Common Properties}
 \begin{itemize}
\item
The network structure is a graph consisting of a number of nodes and a number of edges connecting the nodes.
\item 
Each unit $c$ has an associated position (or a reference vector) $w_c$ in the input space.
\item
Adaption of the reference vectors is done by generating an input signal $\xi$ and moving the reference vector of the nearest or winning units $s_1$ and its direct topological neighbors in the graph towards the input signal:
$$ \triangle w_{s_1} = \epsilon _b(\zeta -w_{s_1}) $$
$$ \triangle w_{i} = \epsilon _n(\zeta -w_{i}) \qquad        (\forall i \in  N_{s_1}) $$
Thereby $N_{s_1}$  denotes the set of direct topological neighbors of $s_1$ (the units which are sharing an edge with $s_1$).
The symbols $\epsilon_b$ and $\epsilon_n$ are adaption constants with $\epsilon_b \gg \epsilon_n$.
\item
At each adaption step local error information is accumulated at the winning unit $s_1$:
$$ \triangle E_{s_1} =  (error\ term) $$

The particular choice of the above error term depends on the application, for example in  vector quantisation one would choose 
$$\triangle E_{s_1} = \left\| w_{s_1} - \zeta\right\|  ^2$$

\item
The accumulated error information is used to determine where to insert new units in the network.
\item
All model parameters are constant over time. 
\end{itemize}
When an insertion is done the error information is locally re-distributed which increases the probability that the next insertion will be somewhere else. The local error variables act as a kind of memory which lasts over several adaption cycles and indicates where much error has occurred.

\subsection{Growing Cell Structures (GCS) }

The purpose of the growing cell structures model is the generation of a topology-preserving mapping from the input space onto a topological structure lower dimensionality~\cite{p3}. This can be seen as a projection onto a non-linear, discretely sampled submanifold. 
This network model can be used to detect clusters of similar patterns according to an unknown probability distribution. The clustering is unsupervised, guided only by sample vectors according to the distribution~\cite{p5}. Currently ongoing studies explore the possibility of constructing three and higher dimensional cell structures.
A simplex (polyhedron with $k$ dimensions, $k+1$ vertices) is used as a basic building block.
 
 GCS network has a topology consisting of many hyper-tetrahedra (triangles in the two-dimensional case), whose vertices (nodes) are associated with weight vectors in the input vector space~\cite{p6}. Each node has a few neighbors, adjacent to it in the topological structure of the network. Periodically, nodes are inserted into and deleted from the GCS network; this occurs at intervals of a fixed number of sample presentations. There are three main aspects of the GCS learning 
algorithm.

\emph{Adaptation:} When an input pattern is presented to the GCS network, a competition takes place, and the weight vectors of each node as well as its immediate topological neighbors are adapted in the direction of the input pattern, albeit with different learning rates. Each node also has an associated signal counter $j$ that estimates the number of input patterns for which that node $j$ was the winner. Counter values also decay with time~\cite{p7}.

\emph{Insertion:} A node is inserted halfway between the node with highest signal counter value and its topologically adjacent neighbor which is at the greatest Euclidean distance. Insertion is followed by the establishment of connections to existing nodes, so that the triangular (or hyper-tetrahedral) network structure continues to be maintained. Voronoi regions and signal counters of adjacent nodes are then adjusted as appropriate.

\emph{Deletion:} Nodes in a GCS with least signal counter value are chosen for deletion. Removal of a node entails removal of the connecting edges, which may leave some nodes ``dangling'', i.e., not part of any triangle (or hyper-tetrahedron in the general case). This eventuality is unthinkable, hence the dangling nodes are also deleted, possibly making other nodes dangle. Repeated elimination of dangling nodes may result in massive purges of the network, erasing the effect of what has been learned earlier, necessitating that the network must rebuild itself. It is possible for subsequent insertions to result in a similar structure, implying that the network may repeatedly shrink and expand, cycling through similar states. In practice, randomness in input presentations may help the algorithm escape such cycles. Fritzke~\cite{p7} has described several examples where the performance of the GCS algorithm is 
superior to that of the SOM algorithm with non-adaptive topology. The deletion mechanism allows GCS to approximate each region accurately, whereas the SOM tends to place several nodes in between these regions, as a consequence of the fixed topological structure.

\subsection{Growing Neural Gas (GNG) }

The Growing Neural Gas (GNG)~\cite{p8} model imposes no explicit constraints on the 
graph. Rather, the graph is generated and continuously updated by competitive Hebbian learning, a technique proposed by Martinetz~\cite{p9}. The core of the 
competitive Hebbian learning method is simply to create an edge between the 
winning and the second winning unit at each adaptation step (if such an edge 
does not already exist). The graph generated is a sub graph of the Delaunay 
triangulation corresponding to the reference vectors~\cite{p10}. The Delaunay triangulation, however,
 is special among all possible triangulations of a point set since it 
has been shown to be optimal for function approximation by Omohundro~\cite{p11}. 
After a fixed number of adaptation steps the unit $q$ with the maximum 
error is determined and a new unit is created between $q$ and one of its neighbors in the graph. Error variables are locally re-distributed and another 
adaptation steps are performed.

The topology of a GNG network reflects the topology of the input signal 
distribution and can have different dimensionalities in different parts of the 
input space. For this reason a visualization is only possible for low-dimensional 
input data. 

\subsection{Self Organising Tree Algorithm  (SOTA)}

SOTA is based both on the Kohonen self-organizing maps  and on the growing cell structures algorithm by Fritzke. The algorithm proposed by Kohonen generates a mapping from a complex input space to a simpler output space. The input space is defined by the experimental input data, whereas the output space consists of a set of nodes arranged according to certain topologies, usually two-dimensional grids. The application of the algorithm produces a reduction in the complexity due to the output space is, usually, smaller than the input space. One of the crucial innovations of SOTA~\cite{p12}  is that the output space has been arranged following a binary tree topology. Additionally, it  incorporates the principles of the growing cell structures algorithm of Fritzke~\cite{p5}  to this binary tree topology. The result has been an algorithm that adapts the number of output nodes arranged in a binary tree to the intrinsic characteristics of the input data set. The growing of the output nodes can be stopped at the desired taxonomic level or, alternatively, they can grow until a complete classification of every sequence in the input data set is reached.

\subsubsection*{\emph{SOTA  features}}
     \begin{itemize}
      \item
      The topology of the network is a binary tree. 
	\item
	Only growing of the network is allowed. 
    \item
     The growing mimics a speciation event, producing two new neurons from the best matching neuron. 
   \item
	Very restricted neighbourhood (problem of asymmetric updating). 
  \item
	Only terminal neurons are directly adapted by the input data, internal neurons are adapted through terminal neurons. 
 \end{itemize}
\subsubsection*{\emph{Algorithm}}
\begin{enumerate}
\item
Initialise system.
\item
Present new input.
\item
Compute distances to all external neurons (tree leaves).
For aligned sequences, the distance between the input sequence $j$ and the neuron $i$ is: 
		$$d_{s_i}{_{c_j}}=\sum_{j=1}^{L}{\frac{1-\sum_{i=1}^{A}s_i[r,l]c_i[r,l] }{L}} $$
where $s_j [{r},{l}]$  is the value for the residue $ r$ of the input sequence node $j$ 
and  $c_i[r,l]$  is the residue $r$ of the neuron $i$.
\item
Select output neuron $i^*$ with minimum distance $d_{ij}$.
\item 
Update neuron $i^*$  and neighbours 
Neurons updated as:
$$ C_i(\tau + 1) = C_i(\tau) + \eta _{\tau,i,j}(s_j-c_i(\tau)) $$
where $\eta _{\tau,i,j}$ is the neighbourhood function for neuron $i$.
\item
If a cycle finished, increase the network size: two new neurons are attached to the neuron with higher resources. This neuron becomes mother neuron and does not receive any more updating.
 
Resources for each terminal neuron $i$ are calculated as an average of the distances of the input sequences assigned to this neuron to itself. 
			$$ R_i = \frac { \sum_{k=1}^{K}d_{{s_k}{c_i}} } {k}  $$
\item
	Repeat by going to Step 2 until convergence (resources are below a threshold).

\end{enumerate}

\subsubsection*{\emph{Advantages}}

Although self-organized neural networks have already been used for the classification problems, using sequence data, the approach presented here is completely different in the sense that a new type of self organizing structure has been developed, which grows accordingly to the hypothetical pattern of speciation which would have given rise to the set of present day sequences analysed. Direct application of Kohonen algorithm~\cite{p2} to data, whose internal relationships are described by means of a binary tree, may produce a correct segregation into the main groups, but lacks a natural way to represent the taxonomic relationships among the individuals. 

Another advantage of the network proposed here is that, since sequences are coded residue by residue, all the information contained in the homologous positions of the alignment is used by the algorithm. 

As can easily be deduced from the description of the algorithm, SOTA convergence depends on the total size of the cells and nodes implied in the phylogeny. From this point of view, time for convergence can be considered to be approximately a linear function of the number of sequences and the number of residues of the data set. This property makes  SOTA a very promising algorithm for the classification of large numbers of sequences, contrarily to other approaches for which the execution time depends on the number of sequences in highly non-linear ways.  Thus execution time is proportional to the cube of the number of sequences and to the number of characters; in least-squares methods, execution time are proportional to the fourth power of the number of sequences and, for algorithms like maximum likelihood, execution times grow exponentially with the number of sequences.

\section { Research Perspectives}

 Growing Neural Networks have been successfully implemented in 
\begin{itemize}

      \item
      Evaluation of High energy Physics experiments~\cite{p14} 
\item
	Computer Graphics (radiosity rendering)~\cite{p8} 
\item
Robust Kinematic learning for a redundant robot arm~\cite{p8} 
\end{itemize}

One promising area where the potential of growing self-organizing networks has not been fully exploited is certainly data mining and knowledge discovery. Clustering huge data sets without knowing in advance the number of clusters is something incremental networks should excel at.

Making hybrid neural networks (combining various self -organizing networks) can result in  efficient clustering, for instance, SOTA is the combination of SOM, Growing Cell Structures (GCS) and the network grows based on the tree data structure. A new neural network can be developed by replacing GCS with Growing Neural Gas (GNG).  As GNG does not impose any constraint on the cell dimension this could be useful for classification purposes.

 Visualisation place an important role in cluster analysis . Advanced Visualisation techniques~\cite{p15} such as Galaxies, Correlation Tools, OmniViz Pro, Hypercube play an important role in analyzing clusters. Integrating these techniques with neural networks can give us  interesting results.

GRB classification~\cite{p16} could be an interesting problem to test with growing neural networks. Possible applications could
be tested on  data sets from the GRB catalogs, for example using
	light curves or band-spectral parameters. 

Separation of gamma from hadrons is one of the important and difficult problems in Gamma-Ray experiments. The classification problem has been addressed with supervised form of neural networks (combination of back propagation and gradient decent back-prorogation). The network separation is based on the study of simulated data~\cite{p13}. It is very likely that severe adjustments have to be made to the simulation to better reflect the data, and the network training has to be redone with the improved simulation. The disadvantage of this approach is the output can be ambiguous and network should be refined constantly for better separation of output. Applying Self-Organizing Networks would be useful as the classification could be  automatic and model-independent.


\begin{thebibliography}{99}

\bibitem{p1} {\tt http://www.batse.msfc.nasa.gov/batse/grb/catalog/current/}

\bibitem{p2}
T.~Kohonen, ``Self-Organizing Maps'', Springer, Berlin (1995).

\bibitem{p3}
B.~Fritzke, ``Growing self-organizing networks - why?'', ESANN, Bruges (1996).

\bibitem{p4}
A.~K.~Jain {\it et al.}, ``Artificial Neural Networks: A tutorial'', IEEE Computing {\bf 29}, 31-44 (1996).

\bibitem{p5}
B.~Fritzke, ``Growing cell structures - a self-organizing network in $k$ dimensions'', ICANN, Brighton (1992).

\bibitem{p6}
B.~Fritzke, ``Kohonen feature maps and growing cell structures - a performance comparison'', NIPS, Denver (1992). 

\bibitem{p7}
B.~Fritzke, ``Unsupervised clustering with growing cell structures'', IJCNN, Seattle (1991). 

\bibitem{p8}
B.~Fritzke, ``Growing self-organizing networks - history, status quo, and perspectives'', in ``Kohonen Maps'',  Proceedings of WSOM-99, eds. E. Oja {\it et al.},  Elsevier (1999).

\bibitem{p9}
T.~M.~Martinez {\it et al.}, ``A neural-gas network learns topologies'', in ``Artificial Neural Networks'', T. Kohonen, K.Makisara, O.Simula, J.Kangas editors,  pages 397-402, North-Holland, Amsterdam (1991).

\bibitem{p10}
S.~Fortune, ``Voronoi Diagrams and Delaunay Triangulations'', in  ``Computing in Euclidean Geometry'', Ding-Zhu Du and Frank Hwang editors, Lecture Notes Series on Computing, volume 1, pages 193-233, World Scientific, Singapore (1992).

\bibitem{p11}
S.~M.~Omohundro, ``The Delaunay Triangulation and Function Learnin'', ICSI Technical Report TR-90-001 (January 1990).

\bibitem{p12}
J.~Dopazo and J. M. Carazo, ``Phylogenetic reconstruction using a growing neural network that adopts the topology of s phylogenetic tree'', J. Mol. Evol. {\bf 44},  226-233 (1997).

\bibitem{p13}
D.~Dorfan {\it et al.}, ``Gamma/Hadron Separation with Neural Networks'', SCIPP 00/31 (2000). 

\bibitem{p14}
M.~Kunze {\it et al.}, ``Growing Cell Structure and Neural Gas - Incremental Neural Networks'', in ``$4^{th}$ Artificial Intelligence in High Energy Physics Workshop'', Pisa (1995).

\bibitem{p15}
{\tt http://www.pnl.gov/infoviz/technologies.html}

\bibitem{p16}
H.~J.~Rajaneimi, P.~Mahonen, ``Classifying GRB using SOM'', APJ {\bf 566}, 202-209 (2001).


\end{thebibliography}
\end{document}